\documentclass{article}

\usepackage{microtype}
\usepackage{graphicx}
\usepackage{subfigure}
\usepackage{booktabs} 

\usepackage{hyperref}

\usepackage{algorithm}
\usepackage{algorithmic}

\usepackage{multirow}

\usepackage[accepted]{icml2023}

\usepackage{dsfont}

\usepackage[capitalize,noabbrev]{cleveref}

\usepackage{enumitem}

\usepackage[textsize=tiny]{todonotes}

\DeclareMathOperator*{\argmin}{arg\,min}

\newcommand{\B}[1]{\boldsymbol{#1}}
\newcommand\norm[1]{\left\lVert#1\right\rVert}

\newcommand{\ours}{SpReME}

\newcommand{\mask}{\B{M}}
\newcommand{\contmask}{\tilde{\B{M}}}

\icmltitlerunning{~\hfill \ours{}: Sparse Regression for Multi-Environment Dynamic Systems \hfill }

\begin{document}

\twocolumn[
\icmltitle{\ours{}: Sparse Regression for Multi-Environment Dynamic Systems}



\icmlsetsymbol{equal}{*}

\begin{icmlauthorlist}
\icmlauthor{MoonJeong Park}{equal,yyy}
\icmlauthor{Youngbin Choi}{equal,yyy}
\icmlauthor{Namhoon Lee}{yyy}
\icmlauthor{Dongwoo Kim}{yyy}
\end{icmlauthorlist}

\icmlaffiliation{yyy}{Graduate School of Artificial Intelligence, POSTECH, Pohang, Korea}
\icmlcorrespondingauthor{Dongwoo Kim}{dongwoo.kim@postech.ac.kr}


\vskip 0.3in
]

\printAffiliationsAndNotice{\icmlEqualContribution} 

\begin{abstract}

Learning dynamical systems is a promising avenue for scientific discoveries.
However, capturing the governing dynamics in multiple environments still remains a challenge:
model-based approaches rely on the fidelity of assumptions made for a single environment, whereas data-driven approaches based on neural networks are often fragile on extrapolating into the future. 
In this work, we develop a method of sparse regression dubbed \ours{} to discover the major dynamics that underlie multiple environments.
Specifically, \ours{} shares a sparse structure of ordinary differential equation (ODE) across different environments in common while allowing each environment to keep the coefficients of ODE terms independently.
We demonstrate that the proposed model captures the correct dynamics from multiple environments over four different dynamic systems with improved prediction performance.
\end{abstract}
\section{Introduction}

Discovering the underlying dynamics of a physical system is an extremely challenging task in many fields of science and engineering;
\textit{e.g.}, it took nearly three decades for experts in fluid mechanics to uncover the true dynamics of flow passing a cylinder \citep{noack2003hierarchy}.
Recently, deep learning-based approaches have gained great attention from the physics and engineering communities with their promises in realistic scenarios including, for instance, those where the physical model is unknown \citep{10.5555/2969239.2969329, pmlr-v80-wang18b}, underlying dynamics are incomplete \citep{APHYNITY} or a form of perturbation from unknown external sources is present \citep{Li2021DisturbanceDF}.

Many current approaches focus on learning a dynamical system from a single environment~\citep{pmlr-v80-long18a}. The approaches with partially known physical models especially show remarkable results in predicting future dynamics~\citep{APHYNITY}. However, learning from a single environment raises the question of whether the learned dynamics can be generalized to the unseen environment. {For example, suppose the training dynamics contain additional perturbation from external sources irrelevant to the governing dynamics.} In that case, the learned model is unlikely to generalize to a new environment where we want to predict future dynamics.

To tackle this issue, some recent works propose learning to capture commonalities across different environments based on neural networks \citep{LEADS, CoDA}.
However, these developments are mostly data-driven approaches without utilizing any prior knowledge of the underlying system.
Hence, they are often challenged on their capability to extrapolate for forecasting future dynamics on unseen time-steps \citep{osti_10277315}.

In this work, we aim to provide a general framework that can uncover governing dynamics from multiple environments with the help of incomplete prior knowledge.
We formulate dynamics discovery as a sparse linear regression problem under the existence of a few major factors of governing dynamics.
Specifically, we define a linear model whose feature functions are the candidates of the governing dynamics. We make the linear models defined for different environments have the same sparse structure through shared binary masks over the feature functions. The shared binary masks determine which feature functions are used to form the linear models.
To uncover the correct sparse structure, we develop an optimization algorithm alternating updates on the shared sparse binary masks and coefficients of features.

In experiments, we compare our model against five baseline models, including model- and data-based approaches. Four different dynamic systems are used to measure the performance of our model. The experimental results show that the proposed model can accurately uncover the underlying dynamics and forecast the evolution of systems over time. Additional analysis reveals that the proposed model performs robust prediction under uncertain environments compared to the other baselines. 

\paragraph{Contributions} The main contributions of this paper can be summarized as follows:
\begin{itemize}[noitemsep, leftmargin=10pt, topsep=0pt]
\item We formulate a dynamic discovery in multiple environments as a sparse linear regression problem.
\item We develop an optimization algorithm that can uncover the true dynamics from multi-environmental observations.
\item We show that the proposed model captures true dynamics on various dynamic systems.
\end{itemize}


\section{Related Work}  

There have been proposed several model-based approaches to leverage some prior knowledge for finding unknown dynamics \citep{APHYNITY, SINDy}.
\citet{APHYNITY} assume that a physical model is partially known and tries to learn unknown parts of its dynamics using neural networks. 
In \citet{SINDy}, candidate elements of unknown dynamics are used as feature (basis) functions to form a sparse regression problem of dynamic systems.
Under the presence of correct basis functions, the underlying dynamics are accurately discovered.  
However, sufficient data from a single environment is needed for using the above methods.

Data-driven approaches have been proposed to uncover the dynamics given the data collected from multiple environments~\citep{LEADS, CoDA, NDP}.
\citet{LEADS} suggests employing two neural network components, one for capturing common dynamics across all environments and the other for capturing environment-specific dynamics.
\citet{CoDA} propose a method that utilizes a single neural network, where the parameters are divided into two parts: common parameters and environment-specific parameters that captures subtle differences across different environments. This approach allows for efficient adaptation to unseen environments, as only the environment-specific parameters need to be re-trained.  
Neural ODE Processes (NDP)~\citep{NDP} combines Neural ODE~\citep{NODE} and Neural Processes~\citep{NP}.
NDP estimates the uncertainty of neural ODE while dealing with multi-environment data. 
The proposed approaches for the multi-environment data are mostly data-driven.
{This raises a question of their ability to forecast the {future state} given the limitation of neural networks on extrapolation~\citep{extrapolation, osti_10277315}.} 


\begin{figure}[t!]
    \centering
    \includegraphics[width=0.99\linewidth]{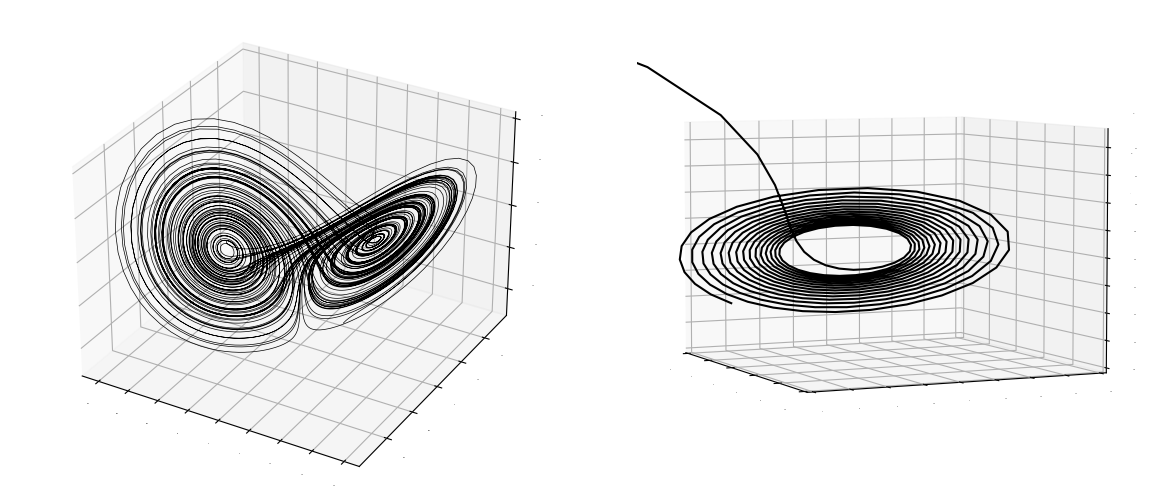}
    \caption{Examples of Lorenz system. Each trajectory shows an evolution of a three-dimensional state over time.}
    \label{fig:Lorenz_example}
\end{figure}

\section{Methodology}

In this section, we first describe a linear model for discovering dynamics for a single environment. We then extend this to the multi-environment and introduce our model to uncover the dynamics from multiple environments. Finally we provide an optimization algorithm to fit the parameters of the model given observations.

\subsection{Linear models for dynamic systems}

Consider a system of dynamics over the continuous time interval between $[0, T]$ whose state $\B{x} = [x_1, \cdots, x_n]^\top \in \mathbb{R}^n$ is defined in an $n$-dimensional space. We consider the dynamics of the form
\begin{equation}\label{eq:LR_form_of_single_env}
    \frac{d\B{x}(t)}{dt}=f(\B{x}(t)),
\end{equation}
where $\B{x}(t)$ denotes the state of dynamics at time $t$. 
A common interest in many science and engineering discipline is to uncover the precise form of function $f$ that governs the observed dynamics. In many cases, the function $f$ consists of a few terms, even if the observed dynamics seem rather complex and chaotic~\citep{broer2010dynamical}. For example, a well-known chaotic dynamic system, the Lorenz system, looks unpredictable as shown in \autoref{fig:Lorenz_example} but can be formalized via the second-order terms on the three-dimensional observations $x,y,z$ as
\begin{equation}
\label{eqn:lorenz}
\bigg(\frac{dx}{dt}, \frac{dy}{dt}, \frac{dz}{dt}\bigg) = \big[\sigma(y-x), x(\rho - z) -y, xy-\beta z \big]^\top,
\end{equation}
where the constants $\sigma, \rho, \beta$ instantiate the real observation.

Given a trajectory, which is a sequence of observation $\B{x}$ over $m$ time steps $t_1, t_2, ..., t_m$, one can formulate the problem of dynamics through the linear model defined as
\begin{equation}
\label{eqn:sindy}
\frac{dx_k(t)}{dt} = \sum_{i=1}^{p} \xi_{ki} \phi_i(\B{x}(t)),
\end{equation}
where $\phi_i:\mathbb{R}^n \rightarrow \mathbb{R}$ is a feature function, c.f., the first and second-order terms in \autoref{eqn:lorenz}, and $\xi_{ki}$ is the coefficient of feature $\phi_i$, c.f, $\gamma, \rho, \beta$ in \autoref{eqn:lorenz}.

Although the approach sounds plausible, it is useless if we do not know which feature functions constitute the model. One common approach is to curate an extensive list of feature functions and fit the model with the sparsity constraints, such as $\ell_1$ regularizer, on $\xi$. In previous work, the sparse regression approach can find the correct dynamics when the list of feature functions contains the true candidate of the underlying dynamics~\citep{SINDy, Kaheman_2020, sindy-var}. When the number of observations is insufficient, however, the model is unlikely to find the true dynamics from the limited observations. 

\subsection{\ours}

We tackle the case where multiple trajectories are observed from various environments, where each environment is defined through its own dynamics of the form in \autoref{eq:LR_form_of_single_env}.
Specifically, consider multiple environments with different dynamics 
\begin{equation}\label{eq:LR_form_of_multi_env}
    \frac{d\B{x}^{(e)}(t)}{dt}=f^{(e)}(\B{x}^{(e)}(t)),
\end{equation}
indexed by environment $e$. Our goal is to find the correct functional form of the true dynamics $f^{(e)}$ with the observations from multiple environments sharing the same functional form, i.e., $f^{(e)}$ consists of the same feature functions $\phi_i$ with varying constants $\xi_i$ across different environments.

One can apply the model in \autoref{eqn:sindy} for each environment to find the true dynamics independently. However, the resulting dynamics may not reveal the same sparse structure across the environments, which is problematic since we know the presence of the common structure. To make the different environments share the same feature functions, we model the dynamics as
\begin{equation}
    \label{eqn:spreme_single}
    \frac{d{x}^{(e)}_k(t)}{dt} = \sum_{i=1}^{p}(M_{ki} \xi_{ki}^{(e)})\phi_i(\B{x}^{(e)}(t)),
\end{equation}
where $M_{ki} \in \{0, 1\}$ is a binary mask. Through the shared binary masks, the model enforces the same sparsity structure across different environments.

The above model can be described compactly over the $n$-dimensional state space in matrix form
\begin{equation}
    \label{eqn:spreme}
    \frac{d\B{x}^{(e)}(t)}{dt} = (\mask\circ\B{\Xi}^{(e)})\Phi(\B{x}^{(e)}(t)),
\end{equation}
where $\Phi(\B{x}^{(e)}(t)) = [\phi_1(\B{x}^{(e)}(t)), \cdots, \phi_p(\B{x}^{(e)}(t))] \in \mathbb{R}^{p}$, $\B{\Xi}^{(e)} \in \mathbb{R}^{n \times p}$ is a collection of $\xi^{(e)}_{ki}$, $\mask\in\{0,1\}^{n\times p}$ is a collection of $M_{ki}$, and $\circ$ indicates element-wise matrix multiplication.

To formulate the model in \autoref{eqn:spreme} as a regression problem, one needs to define a proper loss between the derivative of state ${d\B{x}^{(e)}}/{dt}$ and predicted derivatives. However, the true derivatives are unavailable in general.
To overcome this limitation, we utilize an ordinary differential equation (ODE) solver to use the observed state $\B{x}^{(e)}$ as direct supervision.
The ODE solver takes initial state $\B{x}^{(e)}(t_i)$, estimated functional form of derivatives, initial time $t_i$, and target time $t_j (> t_i)$ as inputs to approximate state at time $t_j$. More precisely, the predicted state at time $T$ is computed as
\begin{align*}
     \hat{\B{x}}^{(e)}(t_j; t_i) &= \B{x}^{(e)}(t_i) + \int_{t_i}^{t_j} \frac{d\B{x}^{(e)}(t)}{dt} dt \\
     &= \text{ODESolve}\left(\B{x}^{(e)}(t_i), \frac{d\B{x}^{(e)}(t)}{dt}, t_i, t_j\right).
\end{align*}
Note that we include $t_i$ in predicted state $\hat{\B{x}}^{(e)}(t_j; t_i)$ to emphasize the time step at the initial state.

With the help of the ODE solver, our objective can be formalized as the following optimization problem:
\begin{align}
\label{eqn:main_objective}
    \argmin_{\mask, \B{\Xi}}\mathcal{L}(\mask, \B{\Xi}) = \sum_e \sum_t \ell(\B{x}^{(e)}(t), \hat{\B{x}}^{(e)}(t)),
\end{align}
subject to a certain sparsity constraint on the mask matrix $\mask$ with a loss function $\ell$.

\subsection{Optimization algorithm}

\begin{algorithm}[t!]
\caption{{Optimization algoritnm}}
\label{alg:algorithm}
    \textbf{Input}: dynamics state {$\{[\B{x}^{(e)}(t_1),\cdots,\B{x}^{(e)}(t_m)]\}_{e=1}^E$}\\
    \textbf{Output}: mask $\B{M}$, coefficient {$\{\B{\Xi}^{(e)}\}_{e=1}^E, \forall e$}
    
\begin{algorithmic}[1] 
    \STATE {Initialize $\B{\Xi}^{(e)}$ via sparse regression in each environment} 
    \STATE {Initialize $\contmask$} \hfill $\triangleright$ \autoref{eqn:mask_initialization}
    \FOR{$\tau=1,\cdots\mathcal{T}$}
        \STATE{Update $\mask$ from $\contmask$ \hfill $\triangleright$	\autoref{eqn:quantize_mask}}
        \STATE {$\B{\Xi}^{(e)} \gets \B{\Xi}^{(e)} - \alpha_\Xi\nabla\mathcal{L}_{\Xi},\forall e$} \hfill $\triangleright$	\autoref{loss_coef_solver}
        \STATE {Prune $\B{\Xi}^{(e)}$ smaller than threshold $\eta_\Xi$}
        \STATE {$\contmask \gets \contmask - \alpha_M\nabla\mathcal{L}_{\contmask}$} \hfill $\triangleright$ \autoref{loss_meta}
        \STATE{Prune $\contmask$ smaller than threshold $\eta_M$ }

    \ENDFOR
    \STATE{Update $\mask$ from $\contmask$ \hfill $\triangleright$	\autoref{eqn:quantize_mask}}
\end{algorithmic}
\end{algorithm}

We provide an optimization algorithm alternating optimization for $\mask$ and $\B{\Xi}$ to solve \autoref{eqn:main_objective}. 

\paragraph{Coefficients \B{$\Xi$} update} Given mask $\mask$, we formulate the regression problem as a minimization of the following loss for coefficient matrices $\{\B{\Xi}^{(e)}\}_{e=1}^{E}$:

\begin{align}\label{loss_coef_solver}
    \mathcal{L}_\Xi(\B{\Xi}) = \sum_e \sum_{i} \ell_\delta \left( \B{x}^{(e)}(t_{i+\eta}),\hat{\B{x}}^{(e)}(t_{i+\eta}; t_i^{'})\right)
\end{align}
where $t_i^{'}$ expresses $t_{\eta\lceil \frac{i}{\eta}\rceil}$, and $\ell_\delta$ is the Huber loss~\citep{10.2307/2238020} with $\delta = 1$.
$\hat{\B{x}}^{(e)}(t_{i+\eta}; t_i^{'})$ is the predicted state at time $t_{i+\eta}$ obtained from an ODE solver using $(\mask\circ\B{\Xi}^{(e)})\Phi(\B{x}^{(e)}(t))$ as a differential equation and $\B{x}^{(e)}(t_i^{'})$ as an initial state, i.e., $\text{ODESolve}(\B{x}^{(e)}(t_{i}), (\mask\circ\B{\Xi}^{(e)})\Phi(\B{x}^{(e)}(t)), t_i^{'}, t_{i+\eta})$.
We compute the gradient by backpropagating the ODE solver directly and use the gradient descent to update $\B{\Xi}^{(e)}$ with step size $\alpha_\Xi$.

To compute the loss, we use $\eta$-step prediction for each state, i.e., $\hat{\B{x}}^{(e)}(t_{i+\eta}; t_i^{'})$ is only predicted from $\B{x}^{(e)}(t_i^{'})$. 
If $\eta$ is one, we perform a one-step prediction for the state where we predict the immediate next time step given a current time step via ODE solver. If $\eta$ is large, we predict the behavior of the state in a long horizon.

\paragraph{Mask $\mask$ update} Given the coefficient of features $\{\B{\Xi}^{(e)}\}_{e=1}^{E}$, optimizing $\mask$ requires enumerating all possible configurations of binary masks, which is intractable in general.
We relax the binary mask to continuous matrix $\contmask\in\mathbb{R}^{p\times n}$ to approximate the discrete mask. The continuous relaxation allows us to use a continuous optimization method.
With the relaxation, we minimize the following objective to update $\contmask$: 
\begin{align}\label{loss_meta}
     \mathcal{L}_M(\contmask) 
    =& \sum_e \sum_{i} \left(\B{x}^{(e)}(t_{i+\eta})-\hat{\B{x}}^{(e)}(t_{i+\eta}; t_i^{'})\right)^2 \notag \\
    & + \lambda(1+s)^\tau\sum_e\norm{\sigma(\contmask)}_1, 
\end{align}
where $\sigma$ is a sigmoid function making the relaxed mask ranges between zero and one, and $\lambda$, $s$, and $\tau$ control the importance of the regularization term. To obtain the predicted state at $i+\eta$-th step ${\hat{\B{x}}^{(e)}(t_{i+\eta}; t_i^{'})}$, we use the integrated state from the initial state $t_i^{'}$ to $t_{i+\eta}$ using an ODE solver, i.e., $\text{ODESolve}(\B{x}^{(e)}(t_i), (\sigma(\contmask)\circ\B{\Xi}^{(e)})\Phi(\B{x}^{(e)}(t)), t_i^{'}, t_{i+\eta})$ as done in the coefficient $\B{\Xi}$ update.
We introduce a scheduler $(1+s)^{\tau}$ for the regularizer weight, where $\tau$ indicates an iteration step.
When $s>0$, the weight of the regularizer increases as the training step increases.
We find that increasing weight helps to explore the sparse structure more in practice.
The relaxed mask is updated with the gradient descent, where the gradient is computed through the ODE solver via back-propagation. 


Once $\contmask$ is optimized, we prune the mask based on the predefined threshold {$\kappa_M$} by setting the corresponding entries to negative infinity to binarize the continuous mask.
We also prune the entries if the corresponding coefficients are zero across all environments since these masks are not updated during optimization.
After pruning, binary mask $\mask$ can be obtained by thresholding: 
\begin{align}\label{eqn:quantize_mask}
    M_{ij} = 
    \begin{cases}
        0, &\text{if}\quad\sigma(\tilde{M}_{ij}) = 0 ,\\
        1, &\text{otherwise,}
    \end{cases}
\end{align}
which is again used to update the coefficient. In practice, we find that further pruning the coefficients based on the predefined threshold {$\kappa_\Xi$} improves the sparsity after the coefficient optimization step.

\paragraph{Initialization}
The use of an ODE solver makes the entire optimization difficult since randomly initialized coefficients may result in a significant loss leading to unstable optimization. 
Therefore, it is necessary to begin with a meaningful initialization in order to alleviate this challenge.

To initialize the coefficient and mask matrix, {we use the sparse regression method of ~\citet{SINDy}, which can be seen as a special case of our model for a single environment without a mask matrix.}
We can directly initialize the coefficients $\B{\Xi}^{(e)}$ from the results of {sparse regression} for each environment with the lowest validation loss. 
To initialize the relaxed mask matrix $\contmask$, we use statistics of the coefficients as follows:
\begin{equation}\label{eqn:mask_initialization}
    \tilde{M}_{ij} = \sigma^{-1}\left(\alpha\frac{\sum_e \mathds{1}_{\Xi_{ij}^{(e)} \neq 0}}{\#\text{ of environments}}\right),
\end{equation}
where $\alpha$, fixed to $0.7$, is introduced to alleviate the gradient vanishing probelm.


The overall optimization algorithm is described in \autoref{alg:algorithm}. Note that the optimization steps for the masking matrix are similar to the straight through estimator~\cite{journals/corr/BengioLC13}. 
Our algorithm alternates the optimization over the continuous mask and the optimization over the coefficients through the quantization.


\paragraph{Validation}
At train time, the ground truth of the mask is not available.
Given the assumption that the learned model extrapolates well on future dynamics, we measure the extrapolation performance as a validation criterion.
We split the given data {$\{[\B{x}^{(e)}(t_1),\cdots,\B{x}^{(e)}(t_m)]\}_{e=1}^E$} into data for timestamp $1$ to $v$ and $v+1$ to $m$.
We use the data point from $1$ to $v$ for training and $v+1$ to $m$ for validation.
The mean-squared error of prediction integrated from time $v+1$ to $m$ without regularization term, i.e., \autoref{loss_meta} with $\lambda=0$, is used as a validation loss. 

\paragraph{Adaptation}
When the trained model encounters a new environment $e^*$, the model needs to adapt to the new environment with partial observations {$[\B{x}^{(e^*)}(t_1),\cdots,\B{x}^{(e^*)}(t_m)]$} available. 
To adapt to the new environment, we only optimize coefficient $\B{\Xi}^{(e^*)}$ since the structure of dynamics is captured in optimized mask $\mask$.
We optimize randomly initialized $\B{\Xi}^{(e^*)}$ using gradient descent with the loss in \autoref{loss_coef_solver} until convergence.
\section{Experimental Setting}

We conduct experiments on four systems of ODE: one linear ordinary differential equation (linear model) and three complex non-linear differential equations (Lotka-Volterra, damped pendulum, and Lorenz models).

\begin{table*}[t!]
\centering
\begin{tabular}{crrrrrrrrrr}
\toprule

\multirow{3}{*}{} & \multicolumn{4}{c}{train}                                                                                                             & \multicolumn{3}{c}{adaptation}                                                            & \multicolumn{3}{c}{test}                                                                                                                                                                                    \\   \cmidrule{2-11} 
                  & \multicolumn{1}{c}{horizon} & \multicolumn{1}{c}{dt} & \multicolumn{1}{c}{\# env.} & \multicolumn{1}{c}{\# traj.}         & \multicolumn{1}{c}{horizon} & \multicolumn{1}{c}{dt} & \multicolumn{1}{c}{\# traj.} & \multicolumn{1}{c}{horizon} & \multicolumn{1}{c}{dt} & \# traj. \\ \midrule
                  
Linear        & \multicolumn{1}{r}{4}                             & \multicolumn{1}{r}{0.05}             & \multicolumn{1}{r}{9}                  & 8                                 & \multicolumn{1}{r}{4}            & \multicolumn{1}{r}{0.05}              & \multicolumn{1}{r}{1}               & \multicolumn{1}{r}{10}           & \multicolumn{1}{r}{0.025}              & 16               \\ 

Lorenz            & \multicolumn{1}{r}{4}                             & \multicolumn{1}{r}{0.05}                    &      9     & 12  & \multicolumn{1}{r}{4}            & \multicolumn{1}{r}{0.05}              & \multicolumn{1}{r}{1}       & \multicolumn{1}{r}{10}           & \multicolumn{1}{r}{0.025}              & 16               \\ 

Lotka Volterra    &  \multicolumn{1}{r}{10}                            & \multicolumn{1}{r}{0.30}                         &     9  & 4       &        \multicolumn{1}{r}{10}           & \multicolumn{1}{r}{0.30}               & \multicolumn{1}{r}{1}                & \multicolumn{1}{r}{25}           & \multicolumn{1}{r}{0.150}              & 32               \\ 

Damped Pendulum   & \multicolumn{1}{r}{4}                             & \multicolumn{1}{r}{0.20}                          &  9    & 8                                & \multicolumn{1}{r}{4}            & \multicolumn{1}{r}{0.20}               & \multicolumn{1}{r}{1}                & \multicolumn{1}{r}{10}           & \multicolumn{1}{r}{0.100}               & 32               \\
\bottomrule
\end{tabular}%
\caption{Description of parameters used to generate trajectories for each ODE system. dt , \# traj. \# env. indicate time intervals, the number of trajectories, and the number of environments, respectively. The training set contains \# of trajectories for each environment. We use shorter time intervals for the test set than those for training and adaptation to measure the interpolation performance.}
\label{table:experiment_setup}
\end{table*}

\begin{table}[]
\resizebox{\columnwidth}{!}{
\begin{tabular}{cccr}
\toprule
\multicolumn{1}{l}{}  & \multicolumn{1}{c}{Parameter} & \multicolumn{1}{c}{Mean} & \multicolumn{1}{c}{Variance} \\ \midrule
Linear                                 &  $\{\alpha, \beta, \gamma, \delta, \omega\}$   & $\{-0.1, 2, -2, -0.1, -0.3\}$ &  0.01     \\
Lorenz                                                     & $\{\sigma, \rho, \beta\}$ &     $\{10, 8/3, 28\}$   &      0.02 \\
DP                                           &    $\{\alpha, \omega_0\}$   &  $\{0.5, 0.98\}$    &     0.01  \\ \bottomrule
\end{tabular}
}
\caption{Means and variances of ODE parameters for each system. We use a normal distribution to sample ODE parameters. DP refers to the damped pendulum. For Lotka-Volterra, we use the same setting used in \citet{CoDA}.}
\label{tab:mean_var}
\end{table}

\subsection{Dynamics}\label{sec:exp-set-dyn}

We describe the details of the four systems of ODE used to measure the performance of dynamic discovery in experiments.

\textbf{Linear}\quad
A linear model is defined via the following ODE:
\begin{equation*}
\frac{d\B{x}}{dt} = \big[\alpha x_0 + \beta x_1, \gamma x_0 + \delta x_1, \omega x_2 \big]^\top
\end{equation*}
where $\alpha, \beta, \gamma, \delta, \omega$ control the dynamics of the system.

\textbf{Lorenz}\quad
A non-linear model representing atmospheric convection with the following dynamics:
\begin{equation*}
\bigg(\frac{dx}{dt}, \frac{dy}{dt}, \frac{dz}{dt}\bigg) = \big[\sigma(y-x), x(\rho - z) -y, xy-\beta z \big]^\top
\end{equation*}
where $\sigma, \rho, \beta > 0$ define the dynamics.
This is a chaotic system where small changes in the initial state produce a drastic difference in relatively short time~\citep{Lorenz}.

\textbf{Lotka-Volterra}\quad
A system for the dynamics of biological interaction between predator and prey can be modeled via~\citep{Lotka}:
\begin{equation*}
\bigg(\frac{du}{dt}, \frac{dv}{dt}\bigg) = \big[\alpha u-\beta uv, \delta uv - \gamma v]^\top
\end{equation*}
where $u, v$ refer to the number of prey and predator, respectively, and  $\alpha, \beta, \gamma, \delta$ control the dynamics of the population.

\textbf{Damped Pendulum}\quad
A model for the dynamics of a damped pendulum~\citep{pendulum} is driven by ODE: 
\begin{equation*}
    \frac{d\theta^2}{dt^2} + \alpha \frac{d\theta}{dt} + \omega_0^2 \sin(\theta) = 0,
\end{equation*}
where damping factor $\alpha$ and initial angular velocity $\omega_0$ defines damping pattern of angle $\theta$.
The state $\B{x}$ is defined as $[\theta, d\theta/dt]^\top$.

\subsection{Dataset Generation}
\label{sec:dataset_gen}

\textbf{Multi-Environment Generation}
To mimic multiple environments, we use various configurations of ODE parameters for each dynamics.
Each set of parameters is sampled from the normal distribution except with Lotka-Volterra, where parameter sets defined in prior works~\cite{CoDA} are used.
We describe the detailed statistics used to generate the environment in \autoref{tab:mean_var}.

\textbf{Trajectory Generation}
To generate trajectories given environment and initial condition, we use LSODA ODE solver~\cite{HindmarshSep2005} to compute the state transition over a finite time horizon with a fixed time interval.
For a given environment, we sample various initial states.
For linear and Lorenz models, the initial state for each trajectory is sampled from the standard normal distribution, i.e., $\mathcal{N}(0, 1)$.
For Lotka-Volterra, each dimension of the initial state is sampled from the unit-mean and unit-variance normal distribution, i.e., $\mathcal{N}(1, 1)$.
For the damped pendulum, the same method used in \citet{APHYNITY} is used for sampling the initial states.

\autoref{table:experiment_setup} describes the detailed statistics for the train, adaptation, and test data sets, including the value of time horizon, time interval, and the number of trajectories for each environment.
The adaptation and test data share the same environment, i.e., the parameters of ODEs are the same, but the trajectories are generated independently with different initial states.
For validation of effective adaptation with few data, we use a single trajectory for the adaptation dataset.

\subsection{Prediction Tasks}

We measure the prediction performance from two different perspectives: in-domain and out-of-domain predictions.
In-domain prediction measures the performance of the trained model in the same environment used for training. In this setting, the additional adaptation to the test set is not performed. Instead, the environment used for training is used to measure the prediction performance.
Out-of-domain prediction measures the performance of the trained model in a different environment from the training. In this setting, we use the adaptation set to optimize the feature coefficient $\B{\Xi}$ and then use the test set to measure the prediction performance. 

For both in-domain and out-of-domain predictions, we predict the state of the trajectories within a training/adaptation time horizon (interpolation) and beyond the training/adaptation time horizon (extrapolation). For the within-time horizon state prediction, we set the interval of test data two times shorter than the training set. By doing so, we can predict the middle points between training states. For the beyond-time horizon prediction, we use two and a half times longer horizon than the training set.

\begin{table*}[t!]
\centering
\begin{tabular}{crrrrrrrr}
\toprule
 & \multicolumn{4}{c}{In-domain}      & \multicolumn{4}{c}{Out-of-domain}     \\   \cmidrule{2-9}
 & \multicolumn{1}{c}{Linear} & \multicolumn{1}{c}{Lorenz} & \multicolumn{1}{c}{LV} & \multicolumn{1}{c}{DP} & \multicolumn{1}{c}{Linear} & \multicolumn{1}{c}{Lorenz} & \multicolumn{1}{c}{LV} & \multicolumn{1}{c}{DP}\\
\midrule

LEADS & 0.4726 & \textbf{4.9767} & \underline{0.0756} & 0.0016& 0.0666 & \underline{41.1628} & \underline{0.0412} & 0.0002  \\
CoDA  & 0.0546 & 26.2325 & 0.2064 & 0.0413 & \underline{0.0012} & 60.2433 & 0.1486 & 0.0012 \\
SINDy & \underline{0.0360} & 70.6965 & - & \underline{2.76e-05}& - & - & - & - \\
Intersection & N/A & N/A & N/A & N/A& 0.0269 & 287.0103 & 0.2559 & \underline{2.7516e-6} \\
Union & N/A & N/A & N/A & N/A & - & 70.1099 & - & - \\
\textbf{SpReME} & \textbf{3.89e-11} & \underline{12.4632} & \textbf{2.90e-06} & \textbf{2.85e-10} & \textbf{9.37e-13} & \textbf{0.1831} & \textbf{7.35e-7} & \textbf{3.76e-11}  \\
\bottomrule

\end{tabular}
\caption{Test MSE in adaptation and in-domain state prediction.
LV and DP refer to Lotka Volterra and Damped Pendulum, respectively.  SINDy is trained with an adaptation dataset. \textbf{Bold} denotes the best result, and \underline{underline} denotes the second best result for each dataset. Numerically unstable results are denoted with a hyphen (-). N/A {means we did not report the performances since the results are the same as those of SINDy.}}
\label{table:main_result1}
\end{table*}

\begin{table}[t!]
\centering
\begin{tabular}{crrrr}
\toprule
\multicolumn{1}{l}{} & \multicolumn{1}{c}{Linear} & \multicolumn{1}{c}{Lorenz} & \multicolumn{1}{c}{LV} & \multicolumn{1}{c}{DP} \\ \midrule
Precision              & 1                              & 0.7                         & 1                       & 1                       \\
Recall                 & 1                              & 1                           & 1                       & 1                       \\
$\#$ candidates ($p$)                 & 56                              & 56                           & 21                       & 24                      \\
\bottomrule
\end{tabular}
\caption{The precision and recall of the trained mask obtained from the lowest validation loss. LV and DP refer to Lotka Volterra and Damped Pendulum, respectively.}
\label{table:mask_result}
\end{table}

\subsection{Baseline Models}
We compare \ours{} with two families of baselines. The first family is based on SINDy~\cite{SINDy}. We use the original SINDy and its two variants,  SINDy-Intersection and SINDy-Union, for handling multi-environment data. The second family is based on neural network approaches, including LEADS~\cite{LEADS} and CoDA~\cite{CoDA}. LEADS and CoDA  are proposed to deal with multi-environment data. For all baseline models, we search hyper-parameters via the validation set. In what follows, we describe the details of each baseline model.


\textbf{SINDy.} SINDy is proposed to uncover the dynamics given a single trajectory from a single environment. To evaluate the performance of SINDy in our multi-environment setting, we train a SINDy in each environment without considering multiple environments. SINDy can use multiple trajectories from a single environment since they only differ in their initial states.

\textbf{SINDy-Intersection.}
SINDy-Intersection is a variant of SINDy for multi-environments. We train a SINDy for each training trajectory independently and then obtain a mask from the learned feature coefficients across all training environments. The entries of the binary mask are set to one if the corresponding coefficients across all environments are non-zero:
\begin{align}\label{eqn:intersection_mask}
    \tilde{M}_{ij} = 
    \begin{cases}
        1, &\text{if}\quad\Xi^{(e)}_{ij} \neq 0 , \quad\forall e \\
        0, &\text{otherwise,}
    \end{cases}
\end{align}
where $\B{\Xi}^{(e)}$ is the coefficient matrix of SINDy for environment $e$.
For out-of-domain prediction, the feature coefficients are trained with our adaptation method.

\begin{table}[t!]
\centering
\resizebox{\linewidth}{!}{%
\begin{tabular}{crrrr}
\toprule
 & \multicolumn{1}{c}{Linear} & \multicolumn{1}{c}{Lorenz} & \multicolumn{1}{c}{LV} & \multicolumn{1}{c}{DP} \\
\midrule

LEADS & 0.0861 & 44.5342 & 0.0668 & 0.0002  \\
CoDA  & 0.0013 & 56.1133 & 0.2262 & 0.0012 \\
SINDy & - & - & - & - \\
Intersection & 0.0345 & 288.3880 & 0.3328 & 1.5800e-06\\
Union & - & 67.6532 & - & - \\
\textbf{SpReME} & \textbf{1.3040 e-12} &\textbf{0.2917} & \textbf{7.6972 e-7} &\textbf{4.1123 e-11} \\
\bottomrule

\end{tabular}
}
\caption{The results of extrapolation in out-of-domain state prediction.
LV and DP refer to Lotka Volterra and Damped Pendulum, respectively.}
\label{table:extrapolation_result}
\end{table}

\textbf{SINDy-Union.}
SINDy-Union is the second variant for the multi-environments, where we set the entry mask to one if any of the coefficients is not zero from the learned SINDy. To be specific, the entries of masks are obtained via
\begin{align}\label{eqn:union_mask}
    \tilde{M}_{ij} = 
    \begin{cases}
        1, &\text{if}\quad\Xi^{(e)}_{ij} \neq 0 , \quad\exists e\\
        0, &\text{otherwise,}
    \end{cases}
\end{align}
where $\B{\Xi}^{(e)}$ is the coefficient matrix of SINDy for environment $e$. The same procedure used for SINDy-Intersection is taken to measure the performance on the test set.

\textbf{LEADS.} LEADS learns dynamics from multi-environments through the combination of the first neural network capturing common
dynamics across all environments and the second neural network capturing the environment-specific dynamics. Specifically, LEADS models the dynamics via
\begin{equation}
    \frac{d\B{x}^{(e)}(t)}{dt} = (f+g^{(e)})(\B{x}^{(e)}(t)),
\end{equation}
where $f$ is a shared neural network across all environments, and $g_e$ is an environment-specific neural network. 
For adaptation in unobserved environment $e^*$, it only trains the new environment-specific neural network $g^{(e^*)}$.

\textbf{CoDA.} CoDA is a meta-learning approach for multi-environments. The method employs a single neural network, denoted as $f$, whose parameters consist of two parts: a set of shared parameters that are common across all environments and a small number of environment-specific parameters:
\begin{equation}
    \frac{d\B{x}^{(e)}(t)}{dt} = f_{\theta+W\phi^{(e)}}(\B{x}^{(e)}(t)),
\end{equation}
where $\theta \in \mathbb{R}^q$ are shared parameters, $\phi^{(e)} \in \mathbb{R}^a$ are environment-specific parameters, and $W \in \mathbb{R}^{q \times a}$ is a matrix for parameter combination. It adapts to the unobserved environment by training environment-specific parameters while keeping the shared parameter fixed.

\begin{figure*}[t!]
    \centering
    \includegraphics[width=0.99\linewidth]{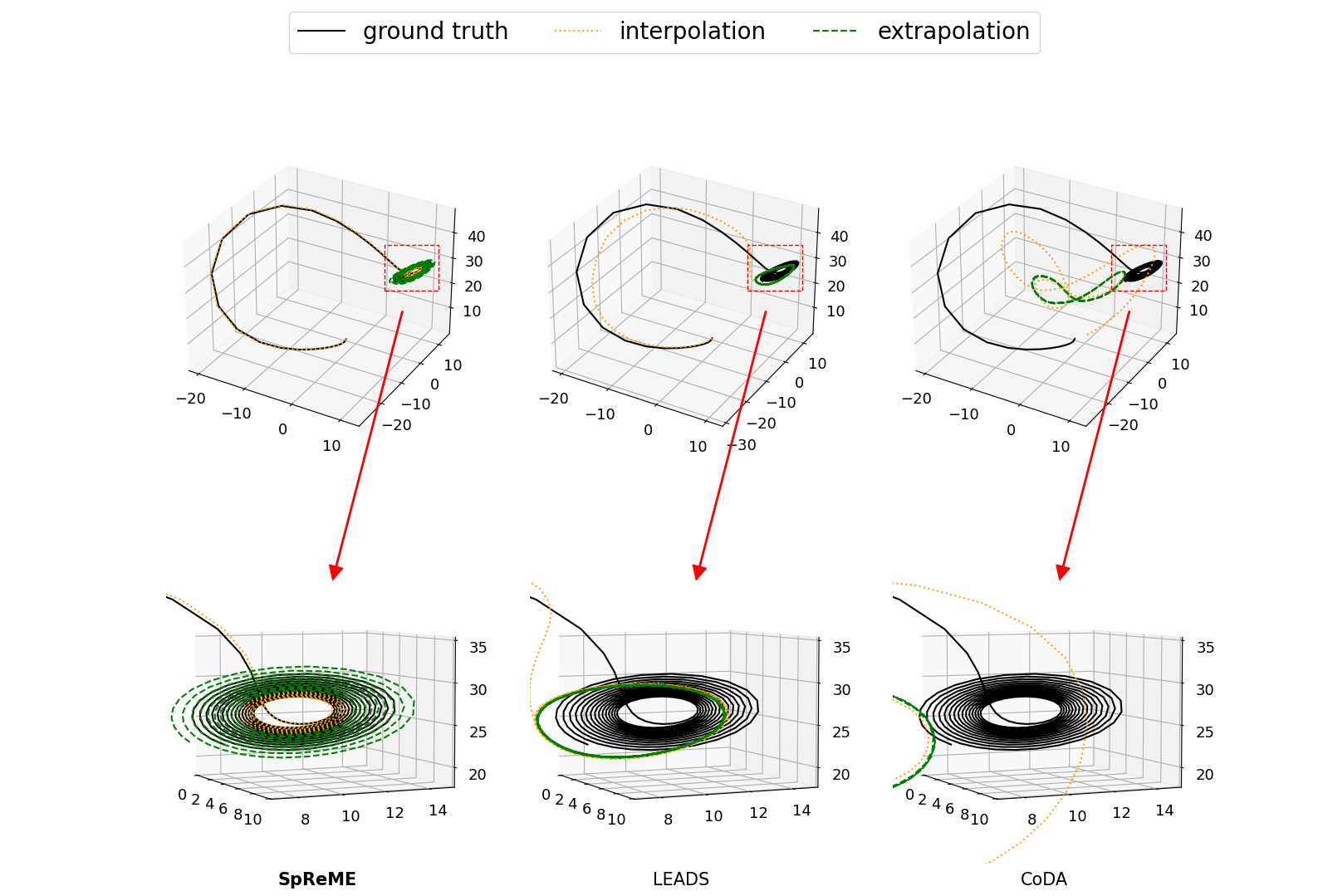}
    \caption{Comparison between the adaptation state prediction of \ours, LEADS and CoDA for Lorenz system. The solid black lines are ground truth, the dotted orange lines are interpolation results, and the dashed green lines are extrapolation results. With the Lorenz model, we provide enlarged views of the extrapolated region (red boxed), changing dramatically from the training trajectory.}
    \label{fig:lorenz_pred}
\end{figure*}

\subsection{Implementation of \ours{}}
{We set $v$ as $0.8m$ for validataion data.}
{To update coefficient $\B{\Xi}$, we set $\eta$ as $1$, and set $\eta$ as $m$ for mask optimization.}
For a successful application of \ours{}, it is important to select the candidate feature functions carefully.
To construct candidate $\Phi(\B{x}^{(e)})$, we use the constant function and polynomial functions with degree five as candidate functions for all dynamics following the settings from \cite{SINDy}.
For the system of the damped pendulum, we additionally use trigonometric feature function $\sin(x_1(t))$, $\sin(x_2(t))$ and $\sin(x_1(t)+x_2(t))$ as candidates.
In all experiments, {we use Adam optimizer~\cite{kingma2014adam}.} The Runge-Kutta 4th order (RK4) solver~\cite{Akinsola23} is used to perform ODESolve.
{Note that RK4 solver is used for all baseline models, and the same candidates $\Phi$ are used for SINDy-based models.}

\section{Experimental Results}


In this section, we provide the experimental results of various prediction tasks. We report the mean-squared error (MSE) as a metric for prediction tasks.

\subsection{Main Results}

\paragraph{In-domain prediction}
The result in \autoref{table:main_result1} shows the in-domain prediction performance of prediction models on the four dynamics. \ours{} performs better than the other baselines except for the Lorenz dataset. The Lorenz dataset is the only case where \ours{} fail to capture the true mask structure. We provide additional details on the accuracy of the binary mask in \autoref{table:mask_result}. 
Note that we did not report the performances of SINDy-Intersection and SINDy-Union since the results are the same as those of SINDy. 

\paragraph{Out-of-domain prediction}
The results of out-of-domain prediction tasks are shown in \autoref{table:main_result1}. \ours{} outperforms the other baseline models over all four dynamics for the out-of-domain prediction. Unlike the results of in-domain prediction, \ours{} outperform the other models on the Lorenz dataset in this case. {\autoref{fig:lorenz_pred} visualizes the result.} Note that SINDy cannot make a stable prediction in this setting since, for the out-of-domain task, SINDy can only use the adaptation set for optimization leading to unstable coefficients of features. The results only focus on the extrapolation is also provided in \autoref{table:extrapolation_result}. For all models, the prediction on extrapolation is generally worse than the overall error reported in \autoref{table:main_result1}. Still, \ours{} outperforms the baselines in this region.

\subsection{Analysis}

\begin{figure}[t!]
    \centering
    \begin{tabular}{cc}
        \multicolumn{2}{c}{\subfigure{\includegraphics[width=0.9\linewidth]{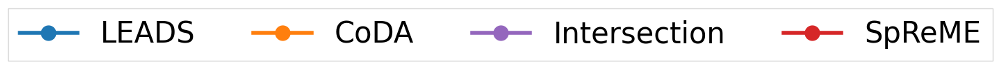}}}\\
        \subfigure{\includegraphics[width=0.45\linewidth ]{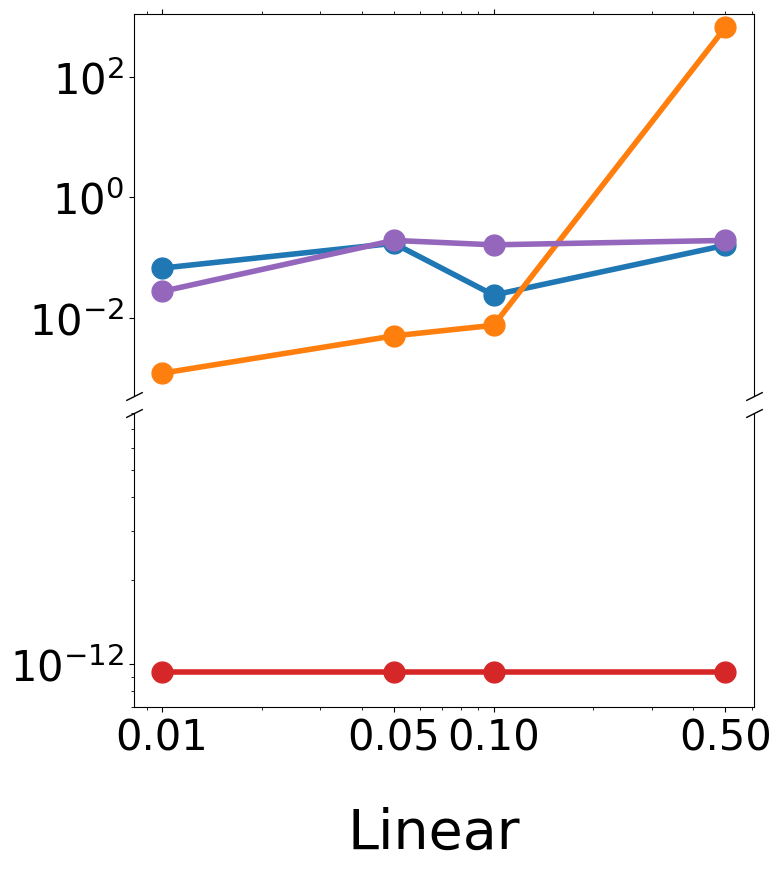}}& 
        \subfigure{\includegraphics[width=0.45\linewidth]{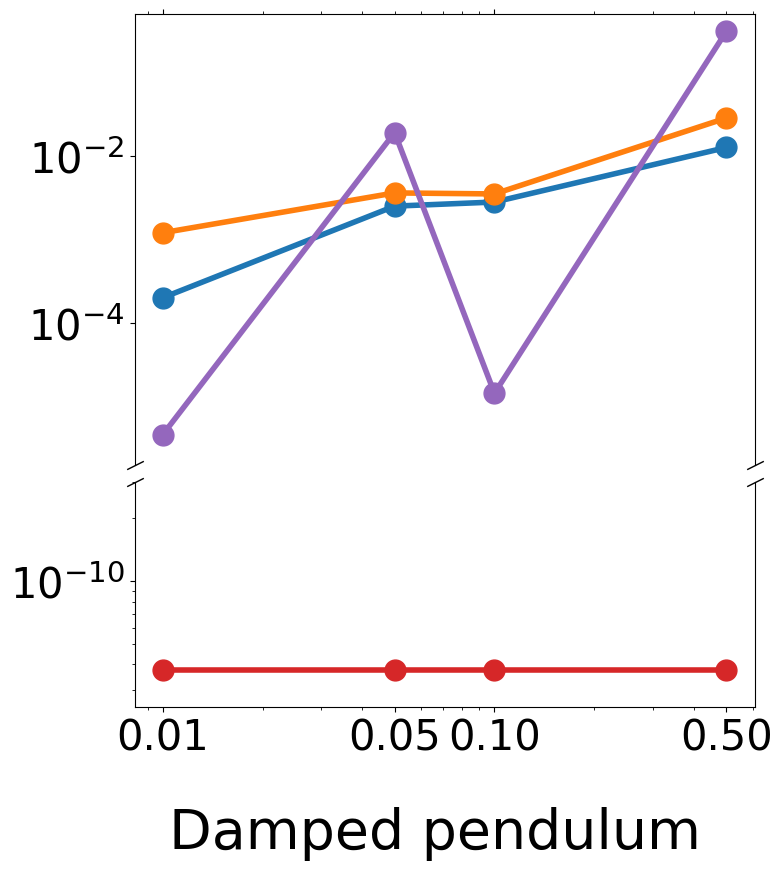}}\\
    \end{tabular}
    \caption{The results of state prediction with increasing uncertainty in training environments. $x$-axis and $y$-axis represent the variance between environments and the MSE performance, respectively. The plots show out-of-domain prediction results on the linear and damped pendulum systems, respectively.}
    \label{fig:main3_var}
\end{figure}

\paragraph{Environmental uncertainty}
To understand how different models cope with environmental uncertainty in optimization, we vary the configuration of the environment by changing the variance of normal distribution, which is used to sample the parameter of ODE. We keep the mean of the normal distribution the same while changing the variance from $0.01$ to $0.5$. \autoref{fig:main3_var} shows the prediction performance on the out-of-domain setting with varying variance in the linear and damped pendulum systems. \ours{} shows stable results even with increasing uncertainty, whereas the other models show increasing error with increasing uncertainty in general.

\begin{figure}[t!]
    \centering
    \begin{tabular}{cc}
        \multicolumn{2}{c}{\subfigure{\includegraphics[width=0.4\linewidth]{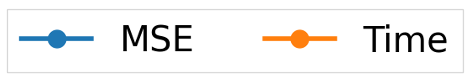}}}\\
        \subfigure{\includegraphics[width=0.45\linewidth ]{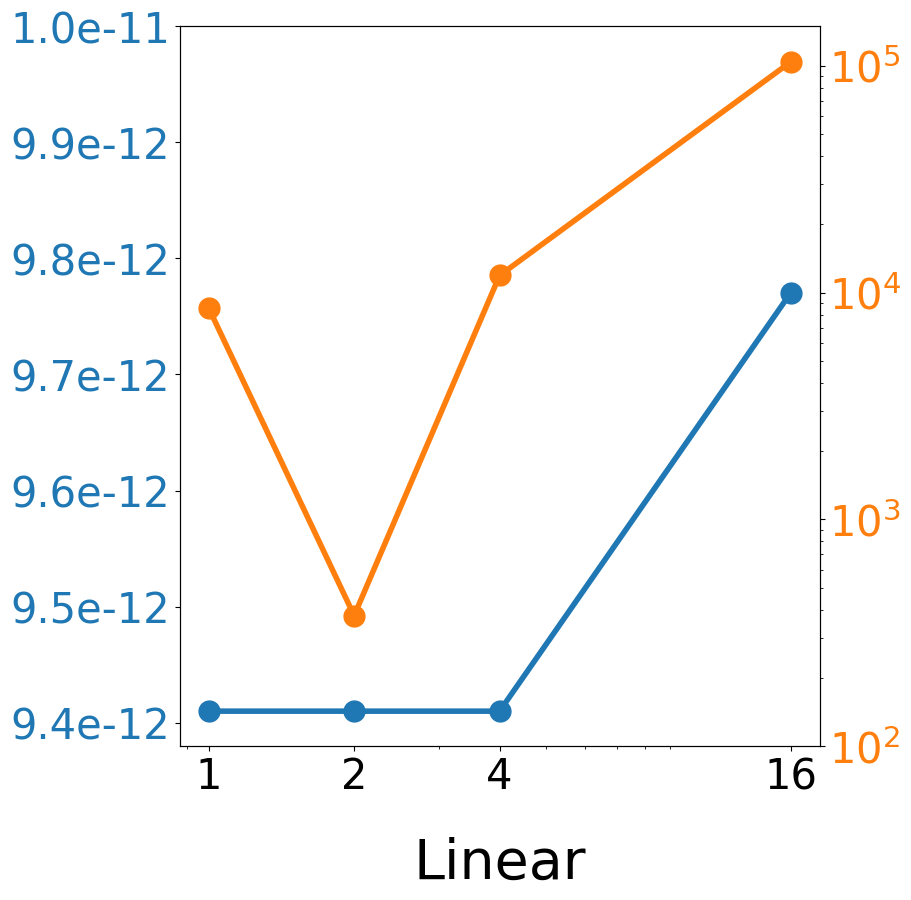}}&  \subfigure{\includegraphics[width=0.45\linewidth ]{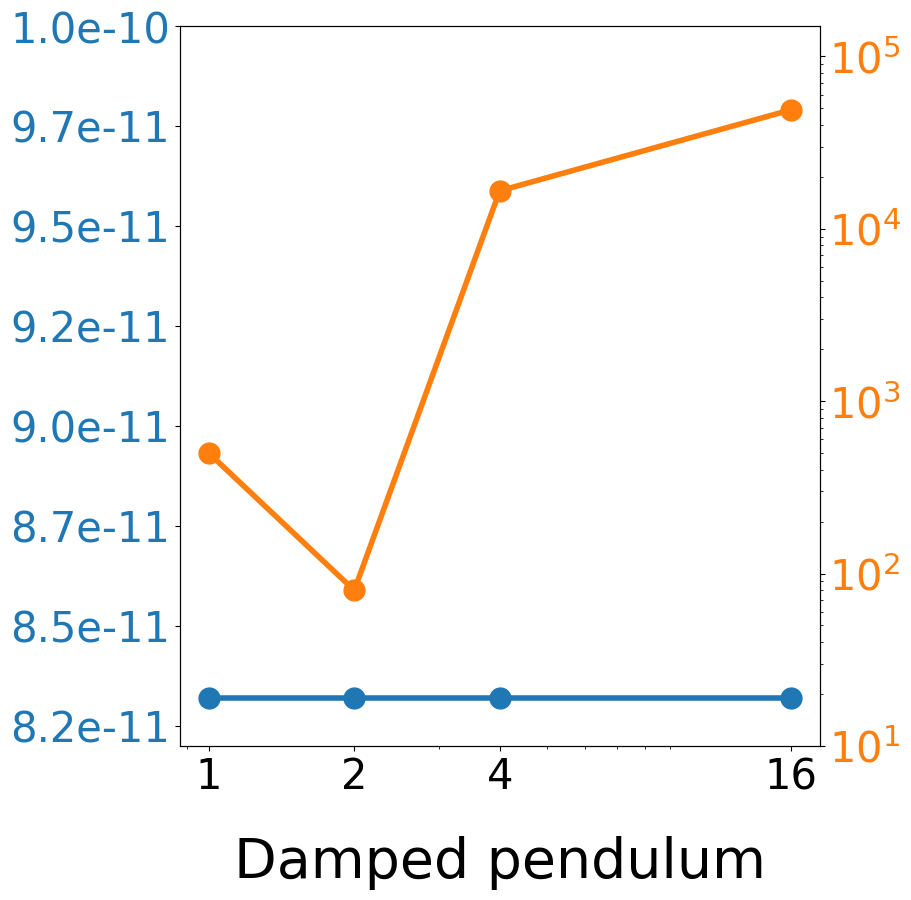}}
    \end{tabular}
    \caption{The result of adaptation performance of models (blue) and time (orange) to learn the mask when the predicted time in mask optimization is varied. $x$-axis represents the predicted time in mask optimization. Experiments are conducted under linear and damped pendulum systems, respectively.}
    \label{fig:longtraj}
\end{figure}

\paragraph{Prediction horizon in optimization}
When optimizing the model parameters, we make a prediction over $\eta$-time steps via an ODE solver and back-propagate over the multiple time steps. We measure the performances in terms of the mean-squared error and computation time with varying $\eta$ and report the results in \autoref{fig:longtraj}. As the results show, a longer prediction horizon may result in a worse performance. We conjecture that the accumulated errors in the ODE solver while integrating over long horizon results in unreliable performance. The computation time is measured until the model parameter is converged. As the results suggest, a proper prediction horizon can achieve prediction performance and efficient computation together.

\section{Conclusion}
We propose \ours{} for discovering dynamics from multiple environments data with incomplete prior knowledge. Unlike prior work, \ours{} use incomplete prior knowledge while dealing with multiple environments. Through the masking scheme, we extract commonalities between different environments. Our model adapts to new environments based on the mask. The experimental results show that \ours{} outperforms LEADS and CoDA in most cases. 

\paragraph{Acknowledgement} This work was partly supported by Institute of Information \& communications Technology Planning \& Evaluation (IITP) grant funded by the Korea government (MSIT) (No.2019-0-01906, Artificial Intelligence Graduate School Program (POSTECH) and No.2022-0-00959, (part2) Few-Shot learning of Causal Inference in Vision and Language for Decision Making) and National Research Foundation of Korea (NRF) grant funded by the Korea government (MSIT) (No.2022R1F1A1064569).


\bibliography{icml2023.bib}

\begin{thebibliography}{25}
\providecommand{\natexlab}[1]{#1}
\providecommand{\url}[1]{\texttt{#1}}
\expandafter\ifx\csname urlstyle\endcsname\relax
  \providecommand{\doi}[1]{doi: #1}\else
  \providecommand{\doi}{doi: \begingroup \urlstyle{rm}\Url}\fi

\bibitem[Akinsola(2023)]{Akinsola23}
Akinsola, V.
\newblock Numerical methods: Euler and runge-kutta.
\newblock In Shah, D.~K. and Ali, D.~A. (eds.), \emph{Qualitative and
  Computational Aspects of Dynamical Systems}, chapter~2. IntechOpen, Rijeka,
  2023.
\newblock \doi{10.5772/intechopen.108533}.
\newblock URL \url{https://doi.org/10.5772/intechopen.108533}.

\bibitem[Baca{\"e}r(2011)]{Lotka}
Baca{\"e}r, N.
\newblock \emph{Lotka, Volterra and the predator--prey system (1920--1926)},
  pp.\  71--76.
\newblock Springer London, London, 2011.
\newblock ISBN 978-0-85729-115-8.
\newblock \doi{10.1007/978-0-85729-115-8_13}.
\newblock URL \url{https://doi.org/10.1007/978-0-85729-115-8_13}.

\bibitem[Bengio et~al.(2013)Bengio, Léonard, and
  Courville]{journals/corr/BengioLC13}
Bengio, Y., Léonard, N., and Courville, A.~C.
\newblock Estimating or propagating gradients through stochastic neurons for
  conditional computation.
\newblock \emph{CoRR}, abs/1308.3432, 2013.
\newblock URL
  \url{http://dblp.uni-trier.de/db/journals/corr/corr1308.html#BengioLC13}.

\bibitem[Broer \& Takens(2010)Broer and Takens]{broer2010dynamical}
Broer, H. and Takens, F.
\newblock \emph{Dynamical Systems and Chaos}.
\newblock Applied Mathematical Sciences. Springer New York, 2010.
\newblock ISBN 9781441968708.
\newblock URL \url{https://books.google.co.kr/books?id=yaov2qvj5YQC}.

\bibitem[Brunton et~al.(2016)Brunton, Proctor, and Kutz]{SINDy}
Brunton, S.~L., Proctor, J.~L., and Kutz, J.~N.
\newblock Discovering governing equations from data by sparse identification of
  nonlinear dynamical systems.
\newblock \emph{PNAS}, 113\penalty0 (15):\penalty0 3932--3937, 2016.
\newblock \doi{10.1073/pnas.1517384113}.
\newblock URL \url{https://www.pnas.org/doi/abs/10.1073/pnas.1517384113}.

\bibitem[Chen et~al.(2018)Chen, Rubanova, Bettencourt, and Duvenaud]{NODE}
Chen, R. T.~Q., Rubanova, Y., Bettencourt, J., and Duvenaud, D.~K.
\newblock Neural ordinary differential equations.
\newblock In \emph{NeurIPS}, 2018.
\newblock URL
  \url{https://proceedings.neurips.cc/paper/2018/file/69386f6bb1dfed68692a24c8686939b9-Paper.pdf}.

\bibitem[Fasel et~al.(2022)Fasel, Kutz, Brunton, and Brunton]{sindy-var}
Fasel, U., Kutz, J., Brunton, B., and Brunton, S.
\newblock Ensemble-sindy: Robust sparse model discovery in the low-data,
  high-noise limit, with active learning and control.
\newblock \emph{Proceedings of the Royal Society A: Mathematical, Physical and
  Engineering Sciences}, 478, 04 2022.
\newblock \doi{10.1098/rspa.2021.0904}.

\bibitem[Garnelo et~al.(2018)Garnelo, Schwarz, Rosenbaum, Viola, Rezende,
  Eslami, and Teh]{NP}
Garnelo, M., Schwarz, J., Rosenbaum, D., Viola, F., Rezende, D.~J., Eslami, S.
  M.~A., and Teh, Y.~W.
\newblock Neural processes.
\newblock \emph{CoRR}, abs/1807.01622, 2018.
\newblock URL \url{http://arxiv.org/abs/1807.01622}.

\bibitem[Hindmarsh \& Petzold(Sep 2005)Hindmarsh and Petzold]{HindmarshSep2005}
Hindmarsh, A.~C. and Petzold, L.~R.
\newblock Lsoda, ordinary differential equation solver for stiff or non-stiff
  system, Sep 2005.
\newblock URL \url{http://inis.iaea.org/search/search.aspx?orig_q=RN:41086668}.

\bibitem[Huber(1964)]{10.2307/2238020}
Huber, P.~J.
\newblock Robust estimation of a location parameter.
\newblock \emph{The Annals of Mathematical Statistics}, 35\penalty0
  (1):\penalty0 73--101, 1964.
\newblock ISSN 00034851.
\newblock URL \url{http://www.jstor.org/stable/2238020}.

\bibitem[Kaheman et~al.(2020)Kaheman, Kutz, and Brunton]{Kaheman_2020}
Kaheman, K., Kutz, J.~N., and Brunton, S.~L.
\newblock {SINDy}-{PI}: a robust algorithm for parallel implicit sparse
  identification of nonlinear dynamics.
\newblock \emph{Proceedings of the Royal Society A: Mathematical, Physical and
  Engineering Sciences}, 476\penalty0 (2242), oct 2020.
\newblock \doi{10.1098/rspa.2020.0279}.
\newblock URL \url{https://doi.org/10.1098%2Frspa.2020.0279}.

\bibitem[Kingma \& Ba(2014)Kingma and Ba]{kingma2014adam}
Kingma, D.~P. and Ba, J.
\newblock Adam: A method for stochastic optimization.
\newblock \emph{arXiv preprint arXiv:1412.6980}, 2014.

\bibitem[Kirchmeyer et~al.(2022)Kirchmeyer, Yin, Dona, Baskiotis,
  Rakotomamonjy, and Gallinari]{CoDA}
Kirchmeyer, M., Yin, Y., Dona, J., Baskiotis, N., Rakotomamonjy, A., and
  Gallinari, P.
\newblock Generalizing to new physical systems via context-informed dynamics
  model.
\newblock In \emph{ICML}, 17--23 Jul 2022.
\newblock URL \url{https://proceedings.mlr.press/v162/kirchmeyer22a.html}.

\bibitem[Li et~al.(2021)Li, Ratliff, and Açikmese]{Li2021DisturbanceDF}
Li, S. H.~Q., Ratliff, L.~J., and Açikmese, B.
\newblock Disturbance decoupling for gradient-based multi-agent learning with
  quadratic costs.
\newblock \emph{L-CSS}, 5:\penalty0 223--228, 2021.

\bibitem[Long et~al.(2018)Long, Lu, Ma, and Dong]{pmlr-v80-long18a}
Long, Z., Lu, Y., Ma, X., and Dong, B.
\newblock {PDE}-net: Learning {PDE}s from data.
\newblock In \emph{ICML}, 10--15 Jul 2018.
\newblock URL \url{https://proceedings.mlr.press/v80/long18a.html}.

\bibitem[Noack et~al.(2003)Noack, Afanasiev, MORZY{\'N}SKI, Tadmor, and
  Thiele]{noack2003hierarchy}
Noack, B.~R., Afanasiev, K., MORZY{\'N}SKI, M., Tadmor, G., and Thiele, F.
\newblock A hierarchy of low-dimensional models for the transient and
  post-transient cylinder wake.
\newblock \emph{JFM}, 497:\penalty0 335--363, 2003.

\bibitem[Norcliffe et~al.(2021)Norcliffe, Bodnar, Day, Moss, and Li{\`o}]{NDP}
Norcliffe, A., Bodnar, C., Day, B., Moss, J., and Li{\`o}, P.
\newblock Neural {ODE} processes.
\newblock In \emph{ICLR}, 2021.
\newblock URL \url{https://openreview.net/forum?id=27acGyyI1BY}.

\bibitem[Quiroga \& Ospina-Henao(2017)Quiroga and Ospina-Henao]{pendulum}
Quiroga, G.~D. and Ospina-Henao, P.~A.
\newblock Dynamics of damped oscillations: physical pendulum.
\newblock \emph{Eur. J. Phys.}, 38\penalty0 (6):\penalty0 065005, oct 2017.
\newblock \doi{10.1088/1361-6404/aa8961}.
\newblock URL \url{https://dx.doi.org/10.1088/1361-6404/aa8961}.

\bibitem[Shi et~al.(2015)Shi, Chen, Wang, Yeung, Wong, and
  Woo]{10.5555/2969239.2969329}
Shi, X., Chen, Z., Wang, H., Yeung, D.-Y., Wong, W.-k., and Woo, W.-c.
\newblock Convolutional lstm network: A machine learning approach for
  precipitation nowcasting.
\newblock In \emph{NeurIPS}, 2015.

\bibitem[Sparrow(1982)]{Lorenz}
Sparrow, C.
\newblock \emph{Introduction and Simple Properties}, pp.\  1--12.
\newblock Springer New York, New York, NY, 1982.
\newblock ISBN 978-1-4612-5767-7.
\newblock \doi{10.1007/978-1-4612-5767-7_1}.
\newblock URL \url{https://doi.org/10.1007/978-1-4612-5767-7_1}.

\bibitem[Wang et~al.(2018)Wang, Gao, Long, Wang, and Yu]{pmlr-v80-wang18b}
Wang, Y., Gao, Z., Long, M., Wang, J., and Yu, P.~S.
\newblock {P}red{RNN}++: Towards a resolution of the deep-in-time dilemma in
  spatiotemporal predictive learning.
\newblock In \emph{ICML}, 10--15 Jul 2018.
\newblock URL \url{https://proceedings.mlr.press/v80/wang18b.html}.

\bibitem[Xu et~al.(2021)Xu, Zhang, Li, Du, Kawarabayashi, and
  Jegelka]{osti_10277315}
Xu, K., Zhang, M., Li, J., Du, S.~S., Kawarabayashi, K.-i., and Jegelka, S.
\newblock How neural networks extrapolate: from feedforward to graph neural
  networks.
\newblock \emph{ICLR}, 2021.
\newblock URL \url{https://par.nsf.gov/biblio/10277315}.

\bibitem[Yin et~al.(2021{\natexlab{a}})Yin, Ayed, de~B{\'{e}}zenac, Baskiotis,
  and Gallinari]{LEADS}
Yin, Y., Ayed, I., de~B{\'{e}}zenac, E., Baskiotis, N., and Gallinari, P.
\newblock {LEADS:} learning dynamical systems that generalize across
  environments.
\newblock \emph{CoRR}, abs/2106.04546, 2021{\natexlab{a}}.
\newblock URL \url{https://arxiv.org/abs/2106.04546}.

\bibitem[Yin et~al.(2021{\natexlab{b}})Yin, GUEN, DONA, de~Bezenac, Ayed,
  THOME, and patrick gallinari]{APHYNITY}
Yin, Y., GUEN, V.~L., DONA, J., de~Bezenac, E., Ayed, I., THOME, N., and
  patrick gallinari.
\newblock Augmenting physical models with deep networks for complex dynamics
  forecasting.
\newblock In \emph{ICLR}, 2021{\natexlab{b}}.
\newblock URL \url{https://openreview.net/forum?id=kmG8vRXTFv}.

\bibitem[Ziyin et~al.(2020)Ziyin, Hartwig, and Ueda]{extrapolation}
Ziyin, L., Hartwig, T., and Ueda, M.
\newblock Neural networks fail to learn periodic functions and how to fix it.
\newblock In \emph{NeurIPS}, 2020.
\newblock URL
  \url{https://proceedings.neurips.cc/paper/2020/file/1160453108d3e537255e9f7b931f4e90-Paper.pdf}.

\end{thebibliography}
\bibliographystyle{icml2023}

\end{document}